\documentclass[11pt]{article}

\usepackage[preprint]{acl}

\usepackage{times}
\usepackage{latexsym}

\usepackage[T1]{fontenc}

\usepackage[utf8]{inputenc}

\usepackage{microtype}

\usepackage{inconsolata}

\usepackage{graphicx}
\usepackage{subcaption}
\usepackage{url}
\usepackage{multirow}

\newcommand{\affilsup}[1]{\rlap{\textsuperscript{\normalfont#1}}}

\title{Simplifying Outcomes of Language Model Component Analyses with ELIA}

\author{
    Aaron Louis Eidt\affilsup{1,2}
    \qquad
    Nils Feldhus\affilsup{1,3}
    \\
    $^1$Technische Universit\"at Berlin \\
    $^2$Fraunhofer Heinrich Hertz Institute \\
    $^3$BIFOLD – Berlin Institute for the Foundations of Learning and Data \\
    {\small \texttt{aaron.eidt@hhi.fraunhofer.de}, \texttt{feldhus@tu-berlin.de} } \\
}

\begin{document}
\maketitle
\begin{abstract}
    While mechanistic interpretability has developed powerful tools to analyze the internal workings of Large Language Models (LLMs), their complexity has created an accessibility gap, limiting their use to specialists. We address this challenge by designing, building, and evaluating ELIA (Explainable Language Interpretability Analysis), an interactive web application that simplifies the outcomes of various language model component analyses for a broader audience. 
    The system integrates three key techniques -- Attribution Analysis, Function Vector Analysis, and Circuit Tracing -- and introduces a novel methodology: using a vision-language model to automatically generate natural language explanations (NLEs) for the complex visualizations produced by these methods. 
    The effectiveness of this approach was empirically validated through a mixed-methods user study, which revealed a clear preference for interactive, explorable interfaces over simpler, static visualizations. 
    A key finding was that the AI-powered explanations helped bridge the knowledge gap for non-experts; a statistical analysis showed no significant correlation between a user's prior LLM experience and their comprehension scores, suggesting that the system reduced barriers to comprehension across experience levels. We conclude that an AI system can indeed simplify complex model analyses, but its true power is unlocked when paired with thoughtful, user-centered design that prioritizes interactivity, specificity, and narrative guidance.
\end{abstract}

\section{Introduction}

\begin{figure}[t]
    \centering
    \includegraphics[width=\columnwidth]{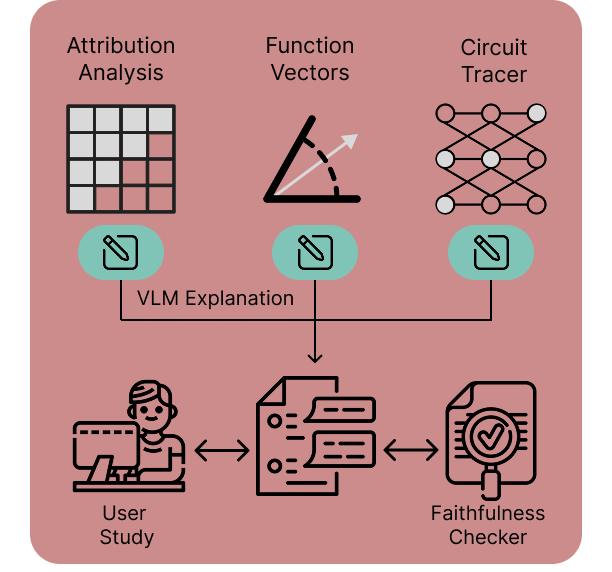}
    \caption{ELIA system overview, including three analysis methods (Attribution Analysis, Function Vectors, and Circuit Tracer) and the explanation generation workflow using VLMs to transform complex interpretability analyses into accessible NLEs. The system is evaluated using a Faithfulness Checker and a user study.}
    \label{fig:elia_architecture}
    \vspace*{-1em}
\end{figure}

The growing capabilities of LLMs are coupled with a proportional increase in their inscrutability. While the field of mechanistic interpretability has made major strides in developing tools to reverse-engineer the internal algorithms of these black-box systems \cite{bereska2024mechanistic, ferrando-2024-primer-inner-workings}, a new challenge has emerged: the outputs of these analyses are often as complex as the models they seek to explain. Techniques such as attribution analysis, which traces predictions to input tokens \cite{sarti-2023-inseq}, or circuit tracing \cite{lindsey2025biology}, which maps specific computational pathways, produce visualizations and data that require specialized expertise to decipher. This creates an accessibility gap, limiting the vital conversation around AI safety and reliability \cite{weidinger-2023-sociotechnical-safety-evaluation} to a small circle of specialists and excluding developers, domain experts, and policymakers who could benefit most.

To bridge this gap, we introduce ELIA (\underline{E}xplainable \underline{L}anguage \underline{I}nterpretability \underline{A}nalysis), an interactive web application\footnote{Demo: \url{https://hf.co/spaces/aaron0eidt/ELIA} \\ GitHub: \url{https://github.com/aaron0eidt/ELIA}} designed to make the outcomes of complex model analyses accessible to a broader audience. ELIA integrates three powerful interpretability techniques, Attribution Analysis \cite{sarti-2023-inseq}, Function Vector Analysis \cite{todd-2024-function-vectors}, and Circuit Tracing \cite{lindsey2025biology}, within a user-centered interface. A vision-language model then generates NLEs for the intricate visualizations produced by these analyses.

Through a mixed-methods user study, we demonstrate the effectiveness of this approach. Our findings show that the AI-generated explanations helped reduce the knowledge gap, enabling non-experts to comprehend complex model behaviors at levels approaching those of users with prior LLM experience. Furthermore, the study revealed a strong user preference for interactive, explorable interfaces over static visualizations. This work provides empirical evidence that the strategic combination of AI-powered explanation and thoughtful, interactive design can significantly lower the barrier to understanding the internal workings of LLMs.


\section{Background and Related Work}

The field of \textbf{NLP interpretability} has progressed through three interconnected streams: moving from correlational to causal analysis, shifting focus from input-output attribution to internal component analysis, and developing methods to communicate these complex findings to a broader audience \cite{saphra-wiegreffe-2024-mechanistic, calderon-reichart-2025-on-behalf-of-the-stakeholders}.

Early interpretability work adapted \textbf{attribution techniques} from computer vision, such as Integrated Gradients \cite{sundararajan2017axiomatic}, to create saliency heatmaps that identify influential input tokens. However, the ``attention is not explanation'' debate and critical sanity checks \cite{jain2019attention, wiegreffe2019attention, adebayo2018sanity} revealed the limitations of these correlational methods, pushing the field toward more rigorous, intervention-based approaches. 

Techniques like activation patching and causal tracing now allow researchers to establish causal links between specific model components and their behavior by \textbf{intervening in the computational process} \citep{zhang2023activationpatching}.
Landmark findings include the identification of induction heads that perform in-context learning \citep{nanda2022induction} and the discovery that entire tasks can be represented by abstract Function Vectors within the model's activation space \citep{todd-2024-function-vectors}. These vectors can be extracted and even composed, demonstrating that models learn structured, portable representations of functions.

Despite these powerful analytical tools, \textbf{communicating the findings remains a significant bottleneck}. The raw outputs, complex graphs, heatmaps, and high-dimensional plots, are often inscrutable to non-experts \cite{colin-2022-what-cannot, schuff-2022-human}. 
To address this, interactive visualization tools like LIT, BertViz, Inseq, and LM Transparency Tool provide explorable interfaces for experts \citep{tenney2020lit, vig2020bertviz, sarti-2023-inseq, tufanov-etal-2024-lmtt}. More recently, the focus has shifted to \textbf{automated explanation} systems that use explainer models to generate natural language descriptions for neuron activity or attention patterns \citep{bills2023language, feldhus-kopf-2025-concept-descriptions}, agents using vision-language models for end-to-end interpretability experiment design \cite{shaham-2024-maia, kim-2025-pursue-agentic-interpretability}, and discovering circuits that represent a particular higher-level function of a model \cite{wang-2023-interpretability-in-the-wild, hanna-2025-circuit-tracer}. However, this automation introduces a critical trade-off between the faithfulness of an explanation (how accurately it reflects the model's process) and its simplicity \cite{feldhus-2023-saliency-map-verbalization, parcalabescu-frank-2024-cc-shap}. Our work is situated at this frontier, aiming to bridge the gap between complex, faithful analyses and simple, accessible explanations through a combination of interactive visualization and AI-generated narrative.

\section{ELIA}

\subsection{System Architecture}

ELIA is an interactive web application designed to make the internal mechanisms of LLMs more transparent and understandable. The system is built using Streamlit\footnote{\url{https://streamlit.io}}, a Python-based framework chosen for its ability to rapidly create data-centric, interactive user interfaces. The backend leverages the scientific Python ecosystem, with PyTorch and the Transformers library for model handling, Plotly\footnote{\url{https://plotly.com}} for dynamic visualizations, and the \textit{inseq} toolkit\footnote{\url{https://inseq.org}} for attribution analyses \citep{sarti-2023-inseq}.

\begin{figure*}[t]
    \centering
    \includegraphics[width=.775\textwidth]{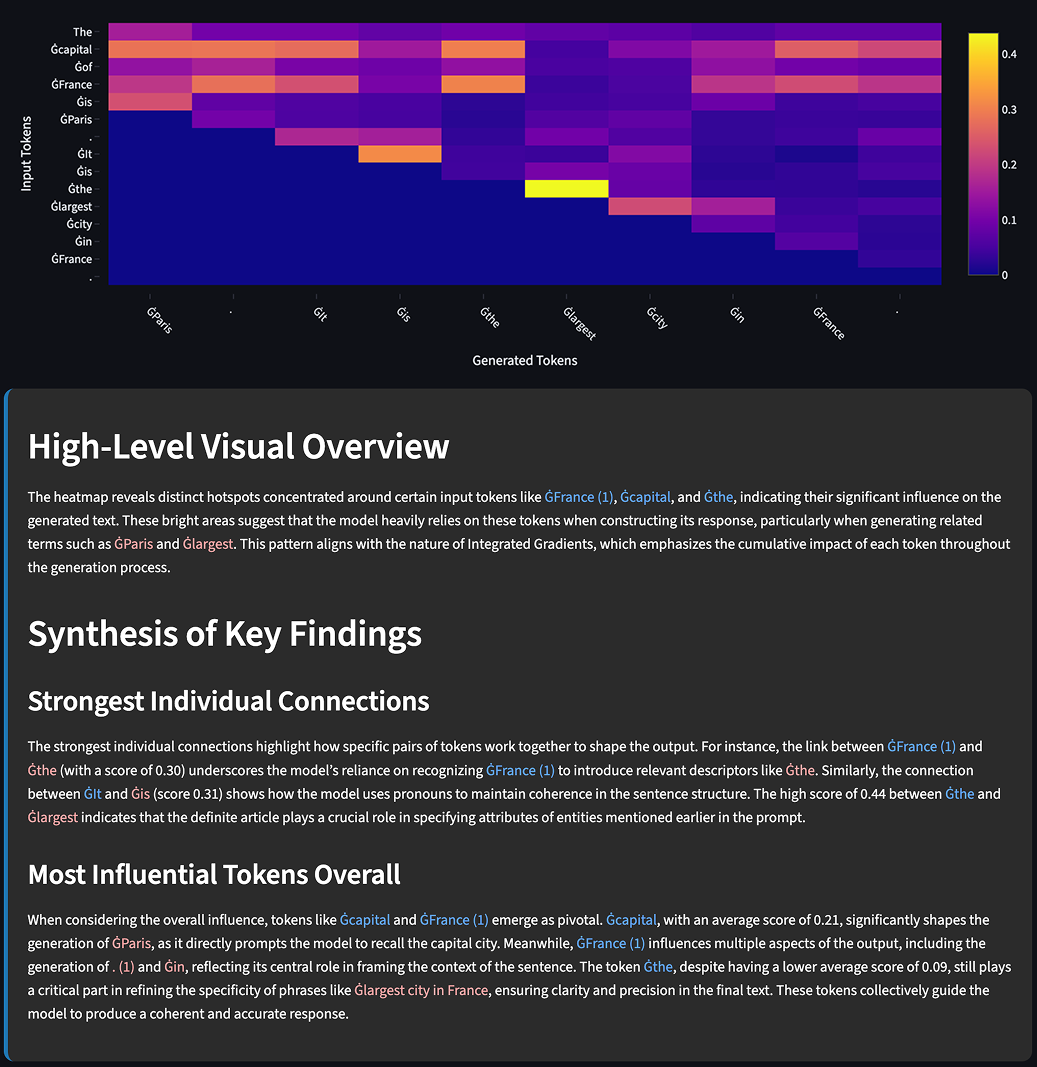}
    \caption{The interactive Attribution Heatmap using Integrated Gradients with an AI-generated natural language explanation. The heatmap visualizes the influence of input tokens on the generated output, and the explanation interprets these results in an accessible narrative.}
    \label{fig:attribution_combined}
    \vspace*{-1em}
\end{figure*}

The architecture is centered around two core models: a \textbf{subject model}, the 7-billion parameter OLMo-2, whose behavior is being analyzed \citep{groeneveld-2024-olmo}; and a vision-enabled \textbf{explanation model} (Qwen2.5-VL-72B) tasked with simplifying the analytical outputs. When a user interacts with one of ELIA's three analysis pages (Attribution Analysis, Function Vector Analysis, or Circuit Tracing), the subject model's internal activations and outputs are visualized (Figure~\ref{fig:elia_architecture}). These visualizations, along with structured textual data, are passed to the explanation model, following prior work on verbalizing explanations \cite{feldhus-2023-saliency-map-verbalization}. This generates a structured, natural language summary of the key insights in an accessible narrative (Figure~\ref{fig:attribution_combined}).

To ensure consistency, API calls to the explanation model are made with a low temperature and a fixed seed, making the generated text largely deterministic. The entire application is internationalized, with full support for both English and German to broaden its accessibility.

\subsection{Faithfulness Verification}

A key component of ELIA's architecture is an automated faithfulness verification system, designed to ensure the reliability of the AI-generated explanations. This system leverages the same explanation model in a multi-step process. First, after generating the initial narrative, the explanation model is prompted again, this time to act as a claim extraction agent, parsing its own text to identify all verifiable, factual statements and structure them as a JSON list, following a similar approach to atomic fact extraction in FActScore \citep{min-etal-2023-factscore}. These claims range from specific quantitative statements (e.g., ``Layer 12 had the highest activation.'') to more abstract semantic assertions (e.g., ``Early layers handle syntax.''). In the second stage, a verification module programmatically checks each claim against the ground-truth data from the underlying analysis. For quantitative claims, this is a direct data comparison. For more abstract semantic claims, the explanation model is called a third time, now tasked to act as a fact-checker to assess the logical plausibility of the claim against the data. The outcome, a \textit{verified} or \textit{contradicted} status for each claim and the supporting evidence, is then presented to the user.

To mitigate the circularity risk of using Qwen2.5-VL-72B for both generation and verification, the verification module operates deterministically (temperature 0.0, fixed seed) and relies heavily on programmatic grounding rather than purely LLM-based judgments. When LLM-based semantic verification is required, the explainer model is constrained by hard-coded rules, negative constraints, and exact synonym-mapping directives, effectively preventing the model from self-affirming its own hallucinations.

\begin{figure*}[t!]
    \centering
    \includegraphics[width=\textwidth]{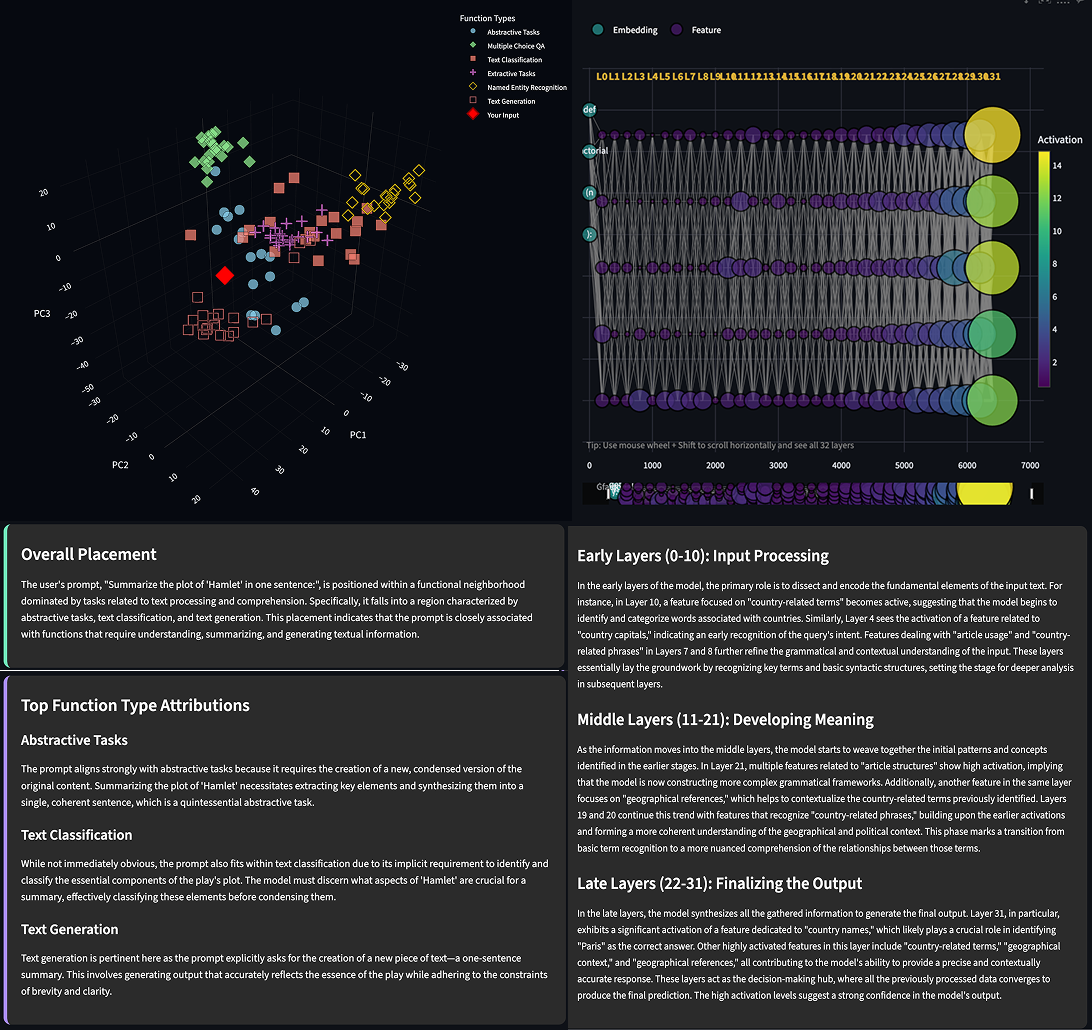}
    \caption{Function Vector and Circuit Trace Analysis visualizations. The 3D PCA plot (left) places the user's prompt in a semantic functional space, while the Circuit Graph (right) traces the flow of information through interpretable features across layers. Both are accompanied by AI-generated explanations.}
    \label{fig:func_circuit_combined}
    \vspace*{-1em}
\end{figure*}

\subsection{Attribution Analysis}
\label{subsec:attribution_analysis}
The Attribution Analysis page provides a granular view of the model's decision-making by quantifying the influence of individual input tokens on the generated output. It integrates two key features: core attribution methods and an influence tracer.

The primary analysis is grounded in three established \textbf{feature attribution} techniques, Saliency, Integrated Gradients, and Occlusion, which are implemented using the \textit{Inseq} toolkit \citep{sarti-2023-inseq}. After the subject model generates text from a user's prompt, the chosen method computes an attribution matrix that is visualized as an interactive heatmap. To translate this complex data into an accessible narrative, the explanation model is given a multi-modal prompt. This prompt combines the heatmap image with a rule-based textual summary that highlights key data points, such as the most influential input tokens and the most affected output tokens, guiding the model to generate a comprehensive, structured explanation of the token-level interactions (Figure~\ref{fig:attribution_combined}). The faithfulness of these explanations is also verified (Figure~\ref{fig:app_attribution_faithfulness} in Appendix).

To further contextualize the model's output, the page includes an \textbf{Influence Tracing} feature that identifies similar documents from the model's training data by performing a $k$-nearest neighbors search against a pre-computed Faiss index of the Dolma dataset, the training corpus for OLMo \citep{douze2024faiss}. The user's prompt is embedded into the same vector space as the training documents, and the most similar examples are retrieved. This allows users to perform a form of data archaeology, exploring potential sources that may have influenced the model's response (Figure~\ref{fig:app_influence_tracer}).

\subsection{Function Vector Analysis}
\label{subsec:function_vector_analysis}
The Function Vector Analysis page offers a high-level, semantic view of the model's behavior. It allows users to explore how the model represents different instructions by comparing a user's prompt to a pre-computed, high-dimensional space of Function Vectors \cite{todd-2024-function-vectors}.

The core of this analysis is a custom dataset of instructional prompts (see Appendix~\ref{sec:appendix_function_vectors} for details), organized into a hierarchy of broad ``function types'' (e.g., abstractive tasks) and specific ``function categories'' (e.g., summarization). The function vector for each category is pre-computed by averaging the final-layer, final-token activations of all its example prompts. When a user enters a new prompt, its own activation vector is computed and compared against this space using cosine similarity.

The results are presented through a suite of interactive visualizations (Figure~\ref{fig:func_circuit_combined}, left). A 3D scatter plot, generated using Principal Component Analysis (PCA), shows the geometric relationship between the user's prompt and the function vector clusters, providing an intuitive map of the model's functional space. This is complemented by a bar chart of the top-scoring function types and a hierarchical sunburst chart that visualizes the similarity scores for all categories. For each of these visualizations, a targeted, AI-powered explanation is generated, synthesizing the key quantitative findings into an accessible, natural-language summary. The faithfulness of the PCA explanation is also verified (see Figure~\ref{fig:app_pca_faithfulness} in the Appendix).

\subsection{Circuit Trace Analysis}
\label{sec:circuit_trace}
This page offers the most granular view of the model's internal workings, building on the circuit tracing framework by \citep{lindsey2025biology}. The analysis is centered around a small autoencoder, called a Cross-Layer Transcoder (CLT) \cite{dunefsky-2024-transcoders}, which is pre-trained to learn a simplified, sparse representation of the OLMo model's internal activations. This CLT is trained on a diverse corpus from the Dolma dataset using $L_1$ sparsity regularization, gradient clipping, and cosine annealing learning rate scheduling (Figure~\ref{fig:app_clt_training} for training dynamics). The CLT is trained to reconstruct the main model's signals while being penalized for using too many features, forcing it to identify the most functionally significant patterns.

These learned features are then given semantic meaning through an automated interpretation step. For each feature, the top-activating input tokens are passed to the explanation model, which generates a concise functional label (e.g., ``identifying JSON syntax''). These interpretable features become the nodes in the main visualization: a layer-by-layer Circuit Graph (Figure~\ref{fig:func_circuit_combined}, right). This graph shows the flow of information from the input tokens, through the activated features, to the final output, with node size and color indicating activation strength and edge thickness representing influence. Additionally, the system provides a view of local path ablations (Figure~\ref{fig:app_pagewise_ablation} in the Appendix), which demonstrates what happens when specific paths in the top feature graph are ablated. 
To make this complex graph accessible, a multi-modal prompt containing the graph image and a summary of key feature activity is used to generate a structured, AI-powered narrative of the information flow. The page also includes interactive "Subnetwork" and "Feature" explorers (see Figures~\ref{fig:app_subnetwork_explorer} and \ref{fig:app_feature_explorer} in the Appendix), allowing users to drill down into the behavior of individual features and their local computational pathways, each augmented with its own targeted, AI-generated explanation. The faithfulness of these explanations is verified and displayed to users (Figure~\ref{fig:app_circuit_faithfulness} in the Appendix).

\section{Faithfulness Analysis}
\label{sec:faithfulness}

Maintaining fidelity between the automatically generated narratives and the underlying model behavior requires dedicated instrumentation on every analysis page. We implement specific verification pipelines that inspect the data powering each visualization, run deterministic checks, and produce aggregate diagnostics that we summarize in Table~\ref{tab:faithfulness_results} and describe in the following for every explanation.

\begin{table}[t]
    \centering
    \small
    \resizebox{\columnwidth}{!}{
    \begin{tabular}{llcc}
    \hline
    \textbf{Component} & \textbf{Feature} & \textbf{Verified} & \textbf{Faith. (\%)} \\
    \hline
    Attribution & Saliency & 59/68 & 86.8\% \\
     & Int. Gradients & 56/61 & 91.8\% \\
     & Occlusion & 66/75 & 88.0\% \\
    \hline
    Func. Vectors & Placement & 23/24 & 95.8\% \\
     & Func. Type & 47/47 & 100.0\% \\
     & Categories & 44/48 & 91.7\% \\
     & Layer Evol. & 36/36 & 100.0\% \\
    \hline
    Circuits & Overview & 45/46 & 97.8\% \\
     & Subnetwork & 132/137 & 96.4\% \\
     & Features & 120/125 & 96.0\% \\
    \hline
    \end{tabular}
    }
    \caption{Faithfulness verification results of each explanation type across all three analysis pages. All explanations are generated by Qwen2.5-VL-72B and verified against ground-truth data from the underlying analyses.}
    \label{tab:faithfulness_results}
    \vspace*{-1em}
\end{table}

\paragraph{Attribution Analysis}
Each time a user runs any attribution method, we collect the raw attribution matrix, compute per-token peak and mean contributions, and identify the strongest interactions between input and output tokens. These statistics drive the automatically generated explanation while also feeding the Faithfulness Checker.

\paragraph{Function Vector Analysis}
The Function Vector workflow exports three independent checkpoints. First, the similarity rankings for function types and categories are recomputed from the cached activations, ensuring that any statement about top matches reflects the actual cosine ordering. Second, the PCA narrative is compared with the cluster centroids that underpin the 3D plot, and a verifier must agree that the textual summary follows from those coordinates. Third, descriptions of layer evolution are cross-checked against the activation norms and change magnitudes from the forward pass.

\paragraph{Circuit Trace Analysis}
Circuit tracing explanations operate over graphs extracted from the Cross-Layer Transcoder. For every narrative, we therefore repeat three stages: we confirm structural statements by inspecting the underlying graph (e.g., upstream/downstream connectivity, active features), validate numeric assertions by reading the stored activation values, and run a semantic check that ensures qualitative summaries remain consistent with the graph evidence. Since the same pipeline is applied to the main circuit view, the feature explorer, and the subnetwork explorer, we obtain uniformly perfect scores for the benchmark prompts.

\begin{figure}[t]
    \centering
    \includegraphics[width=\columnwidth]{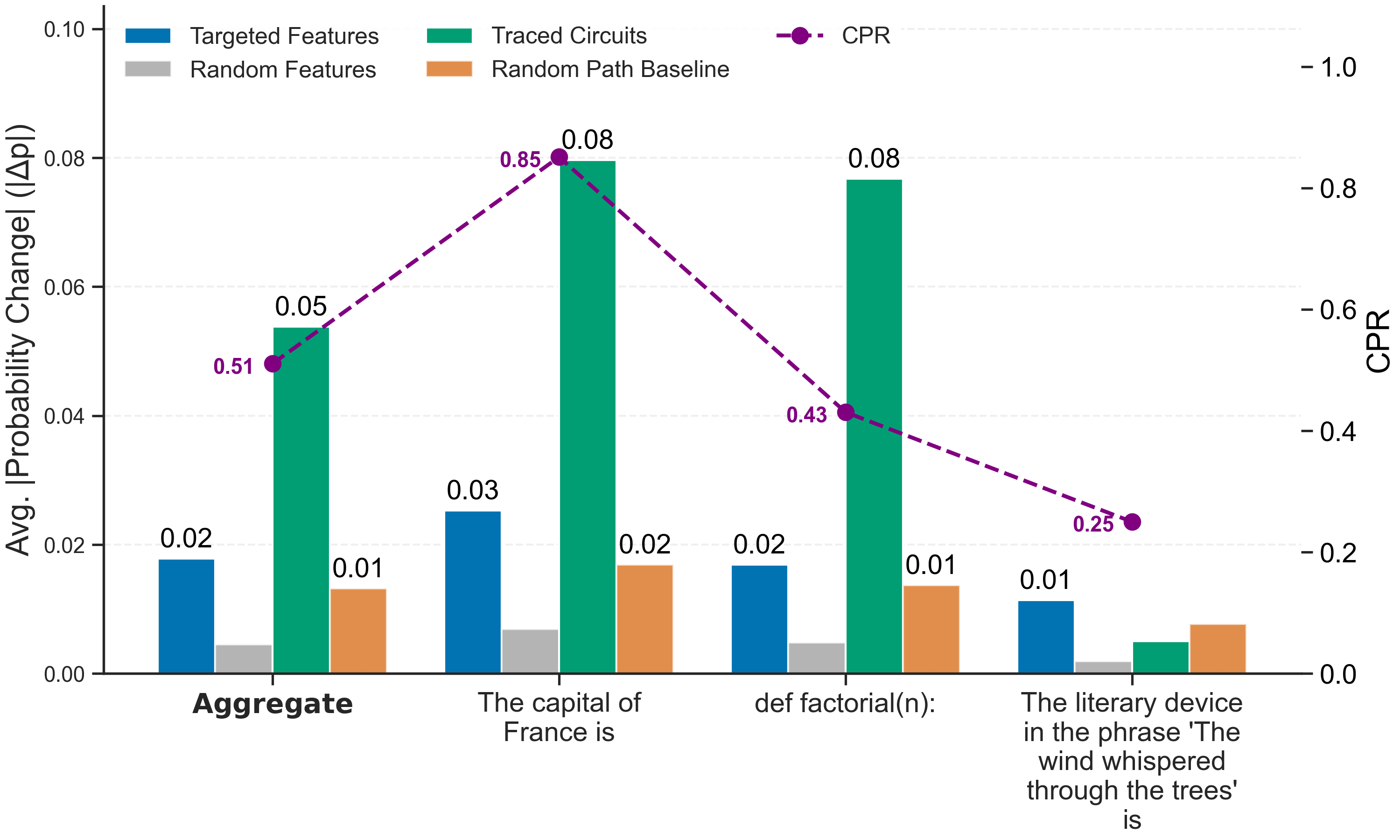}
    \caption{Impact of intervention on model output probability ($|\Delta p|$). We compare the effect of ablating top-$k$ targeted features and traced circuits against random baselines (ablating random features or edges).}
    \label{fig:circuit_metrics}
    \vspace*{-1em}
\end{figure}

To further validate the causal relevance of the discovered circuits, we perform intervention experiments (Figure~\ref{fig:circuit_metrics}). We use a set of exemplary prompts covering knowledge retrieval, code generation, and literary analysis, and by ablating the features and paths identified by the CLT, there is a substantially larger impact on the model's output probability compared to random baselines. This confirms that the system successfully isolates functionally critical components. We also compute the Circuit Performance Ratio (CPR) metric \citep{mueller2025mib}, which quantifies how well a circuit recovers model performance as a function of the fraction of circuit components included.

\section{Use Case: Synergistic Analysis of Knowledge Retrieval}
\label{sec:use_case}

To demonstrate how ELIA's three components can work in concert, we consider a typical knowledge retrieval prompt: ``The capital of France is''.

First, the \textit{Attribution Analysis} identifies the token ``France'' as having the highest saliency score, indicating that the model attends to the subject.

Second, the \textit{Function Vector Analysis} projects the prompt's activation into the pre-computed semantic space, locating it within the ``Abstractive Tasks'' cluster. Specifically, it scores highly on the ``Next Item'' and ``Country Capitals'' categories, suggesting that the model's internal state aligns with the high-level task type of factual completion.

Finally, the \textit{Circuit Trace Analysis} reveals the mechanism. In early layers, features related to ``article usage'' and ``country-related information'' are active, processing the basic syntax and geographical context. In middle layers, features for ``country-related terms'' become prominent, linking the context to specific terminology. In late layers, the model synthesizes this information, with high activation in features related to ``Geographical knowledge'' and ``country-related phrases''. This progression shows a systematic buildup from basic grammar to sophisticated geographical knowledge, explaining how the model generates the answer.

This multi-layered approach allows users to build a more comprehensive understanding by combining evidence from different interpretability methods: establishing input dependence, checking semantic alignment, and inspecting components.

\begin{figure}[t!]
    \centering
    \includegraphics[width=\columnwidth]{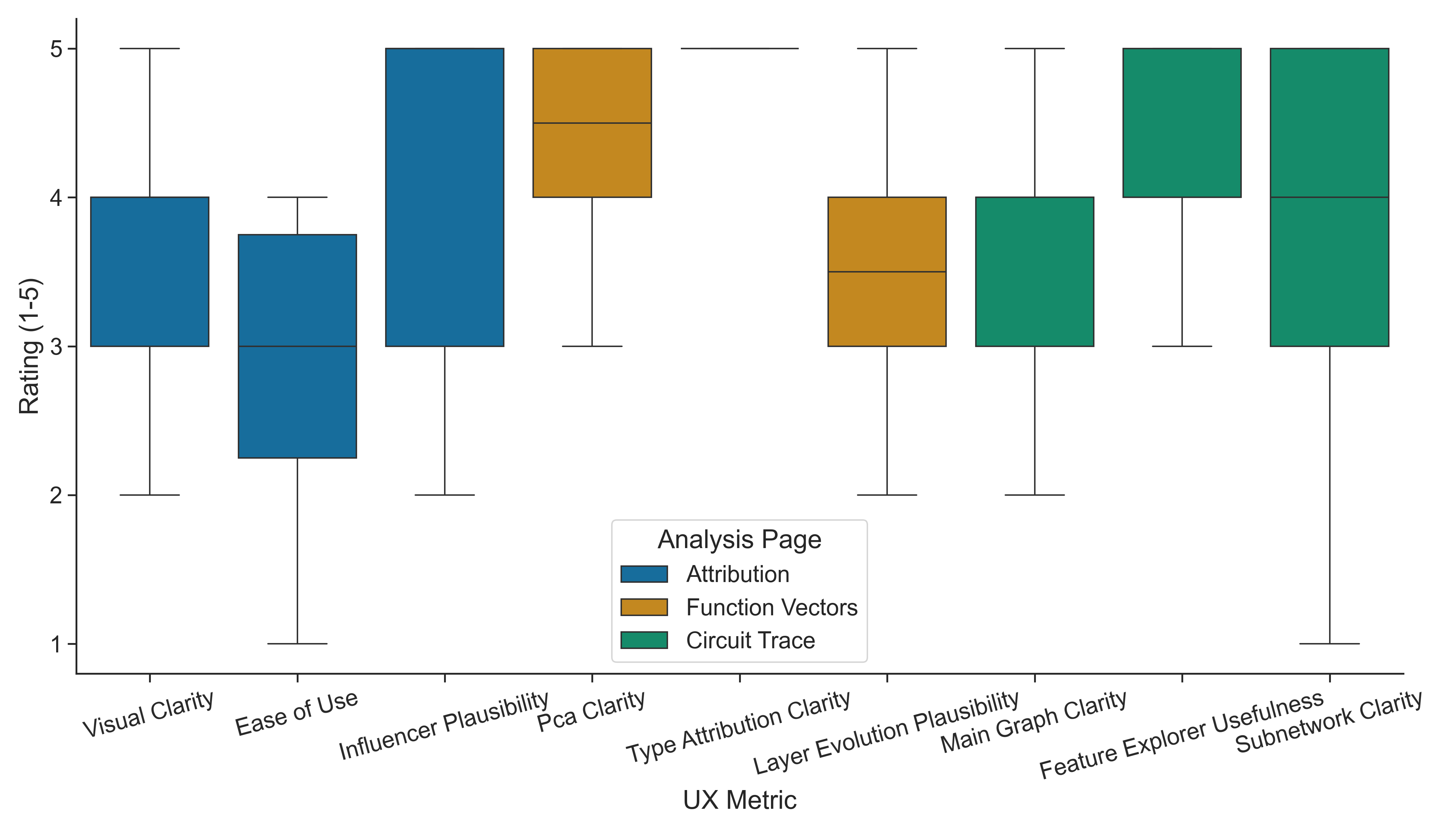}
    \caption{Grouped boxplots of UX ratings for all participants across the three analysis pages.}
    \label{fig:ux_ratings_all}
    \vspace*{-1em}
\end{figure}

\section{User Study}

To empirically evaluate the effectiveness of the explanations and the overall usability of ELIA, a within-subjects user study was conducted with 18 undergraduate computer science students with mostly novice or intermediate experience with LLMs, and designed to assess subjective user experience (UX) and objective comprehension gains. The average completion time was approx. 1h.

\subsection{Quantitative Results}
Participants rated each of the three analysis pages on a 5-point Likert scale according to the following properties: visual clarity, ease of use, plausibility. As shown in Figure \ref{fig:ux_ratings_all}, the Function Vectors and Circuit Trace pages received significantly more positive UX ratings than the Attribution Analysis page (Kruskal-Wallis H-test, $p = 0.006$). The `PCA Clarity` metric on the Function Vectors page received a perfect median score of 5, while the `Ease of Use` for the Attribution Analysis heatmap received the lowest median score of the study (3), indicating it was confusing for users.

To measure objective comprehension, participants answered three multiple-choice questions per page. While there was a positive trend, a Spearman correlation test found no statistically significant relationship between a user's prior LLM experience and their correctness score ($\rho = 0.30, p = 0.23$). The average correctness scores were high across all groups: \textit{Experts} achieved a perfect score of 1.00, \textit{Intermediates} scored 0.98, and even \textit{Novices} reached 0.95. This minimal performance gap suggests that the system helped reduce barriers to comprehension, enabling users with little to no prior knowledge to understand complex model behaviors at levels approaching those of more experienced users.

\subsection{Qualitative Findings}
Analysis of user interview transcripts revealed several key themes. A primary motivation for users was the desire to understand the black box, and the tool was most effective when it provided clear, causal explanations. However, a central challenge was the tension between detail and simplicity; users were often overwhelmed by visualizations that presented too much information at once, such as the main circuit graph.

There is also a clear user preference for interactive visualizations with strong conceptual metaphors. The 3D PCA plot and the Subnetwork Explorer were consistently praised as ``very clear'', while more abstract, static visuals like heatmaps were found to be confusing. Finally, a recurring theme was the need for more integrated, automated guidance. Users frequently relied on the facilitator for context and suggested that adding summaries, tooltips, and adaptive complexity would significantly improve the tool's accessibility.


\section{Conclusion}
ELIA shows that the tools of mechanistic interpretability can become approachable, interactive experiences without giving up analytic depth. By combining the complementary views from attribution heatmaps, function vector projections, and circuit tracing graphs with structured AI narratives and automated faithfulness checks, the platform closes a long-standing accessibility gap: participants in our study reached similar comprehension regardless of prior LLM experience and clearly preferred the explorable interfaces that ELIA provides. The quantitative gains, qualitative feedback, and high verification scores together suggest that interpretability workflows can feel like guided investigations instead of expert-only diagnostics.
We hope to encourage the interpretability community to treat usability and faithfulness as co-equal concerns, nudging the field toward tools that invite participation rather than gatekeep expertise.


\section*{Limitations}
ELIA is currently limited to two languages (English, German), OLMo as the explained model, and the Qwen-VL model as the explainer model.
Richer intervention tools (e.g., counterfactual editing or causal scrubbing) might be necessary to provide a comprehensive user-centric view on interpreting language model behavior.
The user study was limited to subjective ratings and short-term interactions, while studying longer-term usage in professional settings remains a necessary future direction to prove the advantages of ELIA we have so far recorded. 

\section*{Ethical Statement}
The participants in the user study were compensated at or above the minimum wage,
in accordance with the standards of our host institutions' regions.

\section*{Acknowledgments}
We thank the reviewers of the EACL 2026 System Demonstrations track for their valuable feedback. 
This research is funded by the Berlin Institute for the Foundations of Learning and Data (BIFOLD, ref. 01IS18037A)

\bibliography{custom}

\clearpage
\onecolumn
\appendix

\section{Attribution Analysis}
\label{sec:appendix_attribution}

\begin{figure}[h]
    \centering
    \includegraphics[width=0.9\textwidth]{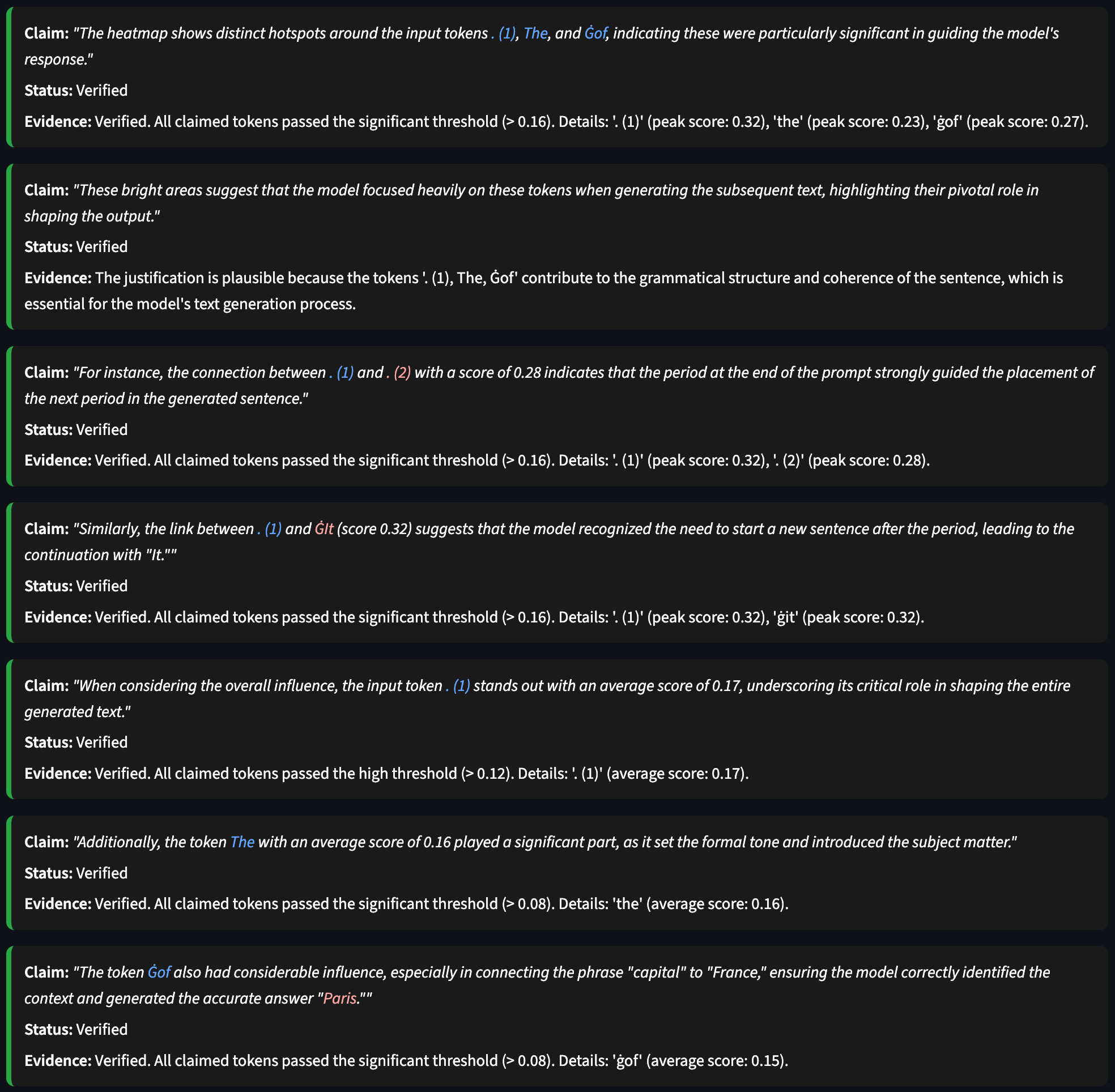}
    \caption{Attribution Analysis: Faithfulness verification of the AI-generated explanation.}
    \label{fig:app_attribution_faithfulness}
\end{figure}

\begin{figure}[h]
    \centering
    \includegraphics[width=0.9\textwidth]{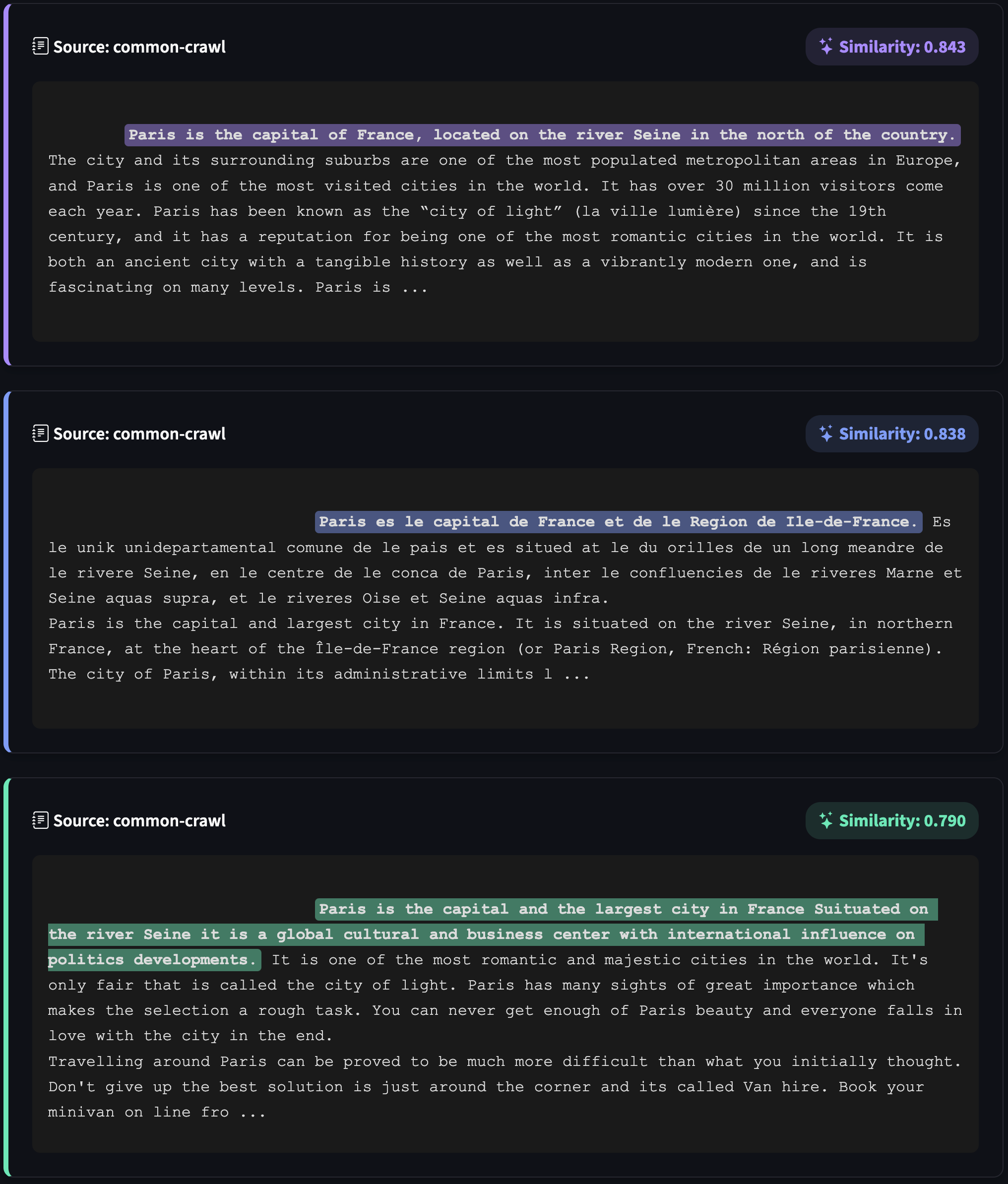}
    \caption{Influence Tracer: Retrieves and displays training documents similar to the prompt.}
    \label{fig:app_influence_tracer}
\end{figure}

\clearpage
\section{Function Vector Analysis}
\label{sec:appendix_function_vectors}
The Function Vector space is constructed using a custom bilingual dataset (English and German) to ensure robust, cross-lingual functional mappings. For each category, we pass the instructional prompts through OLMo-2-1124-7B and extract the activation vector from the final token position of the last hidden layer. The definitive function vector for a given category is computed as the mean of these activations. To maintain interactivity and minimize computational overhead in the live application, this high-dimensional space is pre-computed offline. During real-time usage, analyzing a new prompt requires only a single forward pass to extract its activation vector, followed by a highly efficient cosine similarity comparison against the pre-cached semantic space.

\begin{table*}[!h]
    \centering
    \footnotesize
    \setlength{\tabcolsep}{3em}
    \resizebox{\textwidth}{!}{
        \begin{tabular}{lp{2.2cm}p{7.2cm}}
        \hline
        \textbf{Function Type} & \textbf{Category} & \textbf{Example Prompts} \\
        \hline
        \multirow{2}{*}{Abstractive Tasks} & \texttt{country\_capital} & The capital of France is \\
         & & What city serves as the capital of Japan? \\
        \cline{2-3}
         & \texttt{translation\_german} & Translating 'hello' into German gives \\
         & & What would a German speaker say for 'world'? \\
        \cline{2-3}
         & \texttt{next\_item} & After 'Monday' comes \\
         & & The number following 5 is \\
        \hline
        \multirow{2}{*}{Multiple Choice QA} & \texttt{commonsense\_qa} & What happens when you mix red and blue? \\
         & & Why do people wear coats in winter? \\
        \cline{2-3}
         & \texttt{math\_qa} & What is 15 multiplied by 8? \\
         & & Calculate the area of a square with side 5 \\
        \cline{2-3}
         & \texttt{geography\_qa} & Which is the largest ocean? \\
         & & What is the longest river in the world? \\
        \hline
        \multirow{2}{*}{Text Classification} & \texttt{sentiment\_analysis} & Is this text positive or negative? \\
         & & What emotion does this express? \\
        \cline{2-3}
         & \texttt{language\_detection} & What language is this text written in? \\
         & & Identify the language of this sentence \\
        \cline{2-3}
         & \texttt{spam\_detection} & Is this message spam? \\
         & & Classify this email as spam or legitimate \\
        \hline
        \multirow{2}{*}{Extractive Tasks} & \texttt{adjective\_vs\_verb} & Is 'running' an adjective or verb? \\
         & & Classify 'beautiful' as adjective or verb \\
        \cline{2-3}
         & \texttt{living\_vs\_nonliving} & Is 'tree' living or non-living? \\
         & & Classify 'car' as living or non-living \\
        \cline{2-3}
         & \texttt{concrete\_vs\_abstract} & Is 'happiness' concrete or abstract? \\
         & & Classify 'table' as concrete or abstract \\
        \hline
        \multirow{2}{*}{Named Entity Recognition} & \texttt{ner\_person} & Identify the person name in this text \\
         & & Extract all person names mentioned \\
        \cline{2-3}
         & \texttt{ner\_location} & What location is mentioned here? \\
         & & Extract all place names from the text \\
        \cline{2-3}
         & \texttt{ner\_organization} & Find the organization name \\
         & & Extract company or institution names \\
        \hline
        \multirow{2}{*}{Text Generation} & \texttt{complete\_sentence} & Complete this sentence: "The weather today is" \\
         & & Finish the thought: "In the future, we will" \\
        \cline{2-3}
         & \texttt{continue\_story} & Continue the story: "Once upon a time..." \\
         & & What happens next in this story? \\
        \cline{2-3}
         & \texttt{question\_generation} & Generate a question about this topic \\
         & & Create a question based on this text \\
        \hline
    \end{tabular}
    }
    \caption{Overview of the Function Vector dataset structure. The dataset contains 6 function types with 120 total categories. Each category includes 5 example prompts. This table shows representative examples from each function type.}
    \label{tab:function_vector_dataset}
\end{table*}

\begin{figure*}[h]
    \centering
    \includegraphics[width=0.9\textwidth]{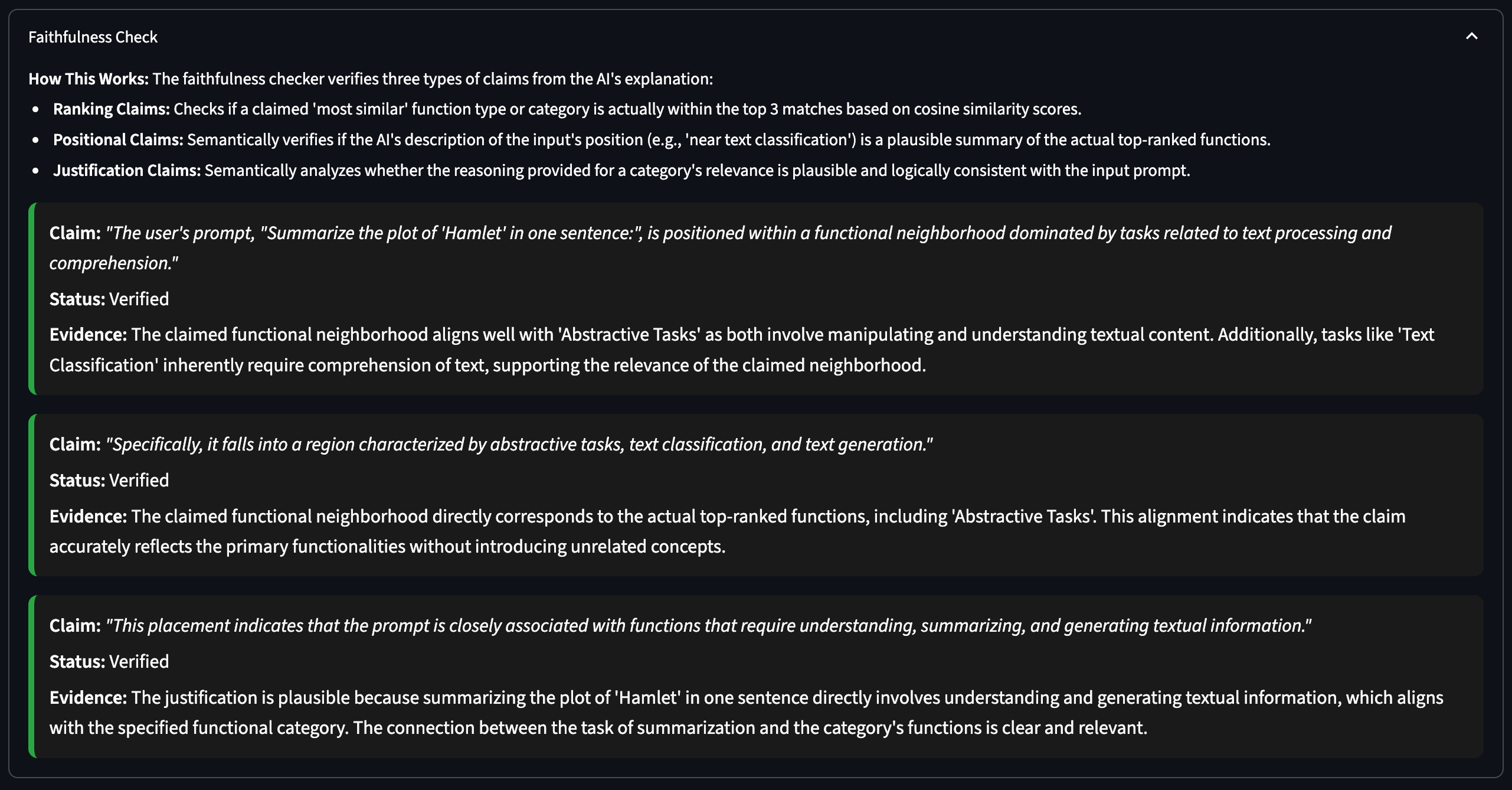}
    \caption{Function Vector Analysis: Faithfulness verification of the AI-generated PCA explanation.}
    \label{fig:app_pca_faithfulness}
\end{figure*}

\clearpage

\section{Circuit Trace Analysis}
\label{sec:appendix_circuit}
\begin{figure}[h!]
    \centering
    \includegraphics[width=\textwidth]{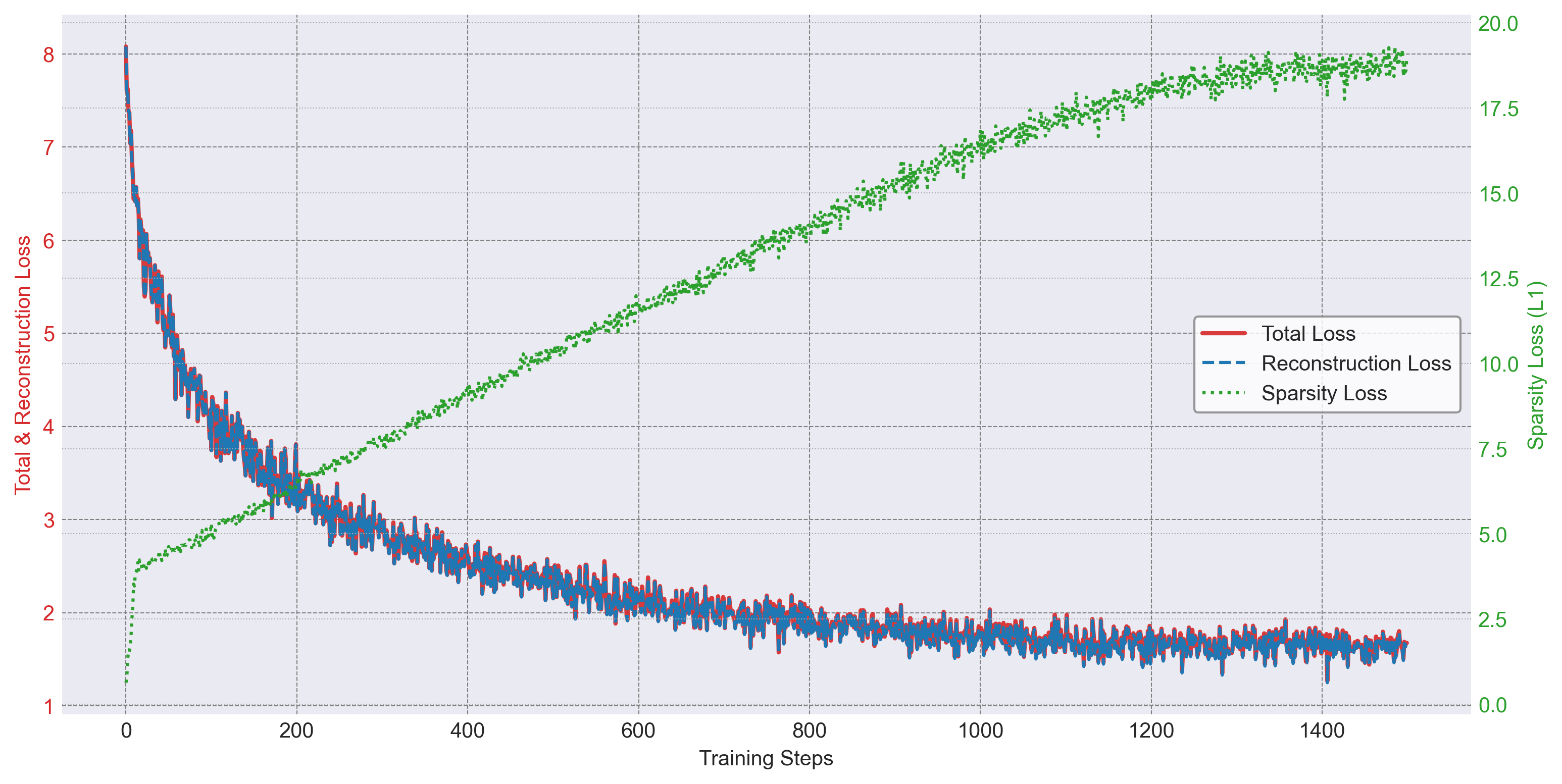}
    \caption{Cross-Layer Transcoder training dynamics showing Total Loss, Reconstruction Loss, and Sparsity Loss ($L_1$) over 1500 training steps. The CLT is trained on the Dolma dataset with $L_1$ sparsity regularization, gradient clipping, and cosine annealing learning rate scheduling.}
    \label{fig:app_clt_training}
\end{figure}

The Cross-Layer Transcoder (CLT) is trained offline to learn a sparse, simplified representation of the model's internal activations. The model was trained on text samples from the Dolma dataset using a batch size of 16 over 1,500 training steps. 
Our architecture maps the hidden dimension to 512 interpretable features per layer, utilizing a JumpReLU activation (threshold = 0.0) alongside an $L_1$ sparsity penalty ($\lambda = 1e^{-3}$). Optimization was performed using Adam (learning rate: $3e^{-4}$) with cosine annealing and gradient clipping (max norm: 1.0) to ensure stable convergence. The training dynamics are visualized in Figure~\ref{fig:app_clt_training}.

By performing this resource-intensive training offline, the live computational cost of ELIA is drastically reduced. Real-time operations are limited to standard forward passes and asynchronous API calls for the natural language explanations, ensuring the interface remains highly interactive without requiring extensive local compute resources.

\begin{figure}[h!]
    \centering
    \includegraphics[width=\textwidth]{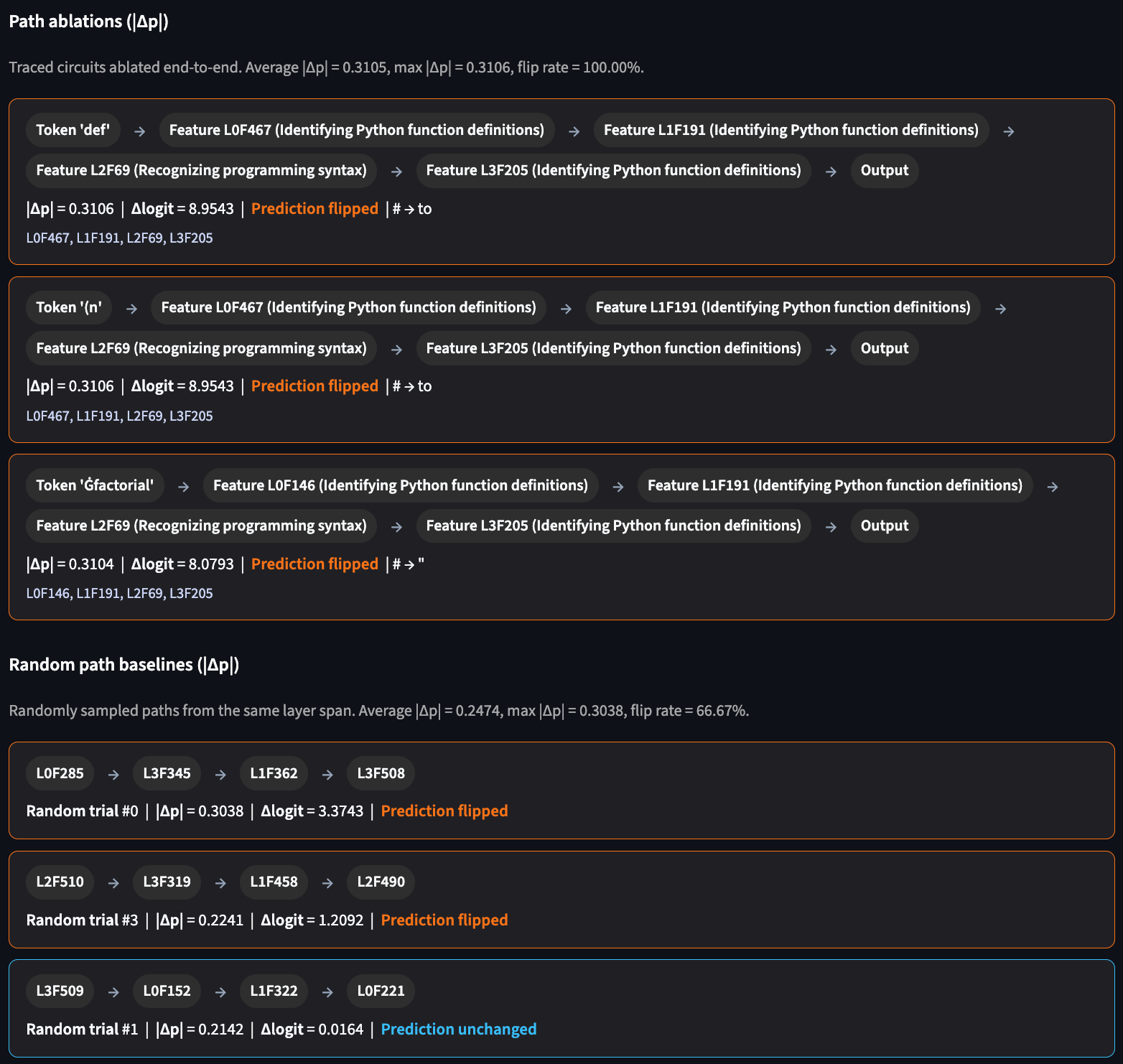}
    \caption{Circuit Trace Analysis: Local path ablations. This view shows the effect of ablating specific paths within the top feature graph shown in the main circuit trace visualization.}
    \label{fig:app_pagewise_ablation}
\end{figure}

\begin{figure*}[h]
    \centering
    \includegraphics[width=\textwidth]{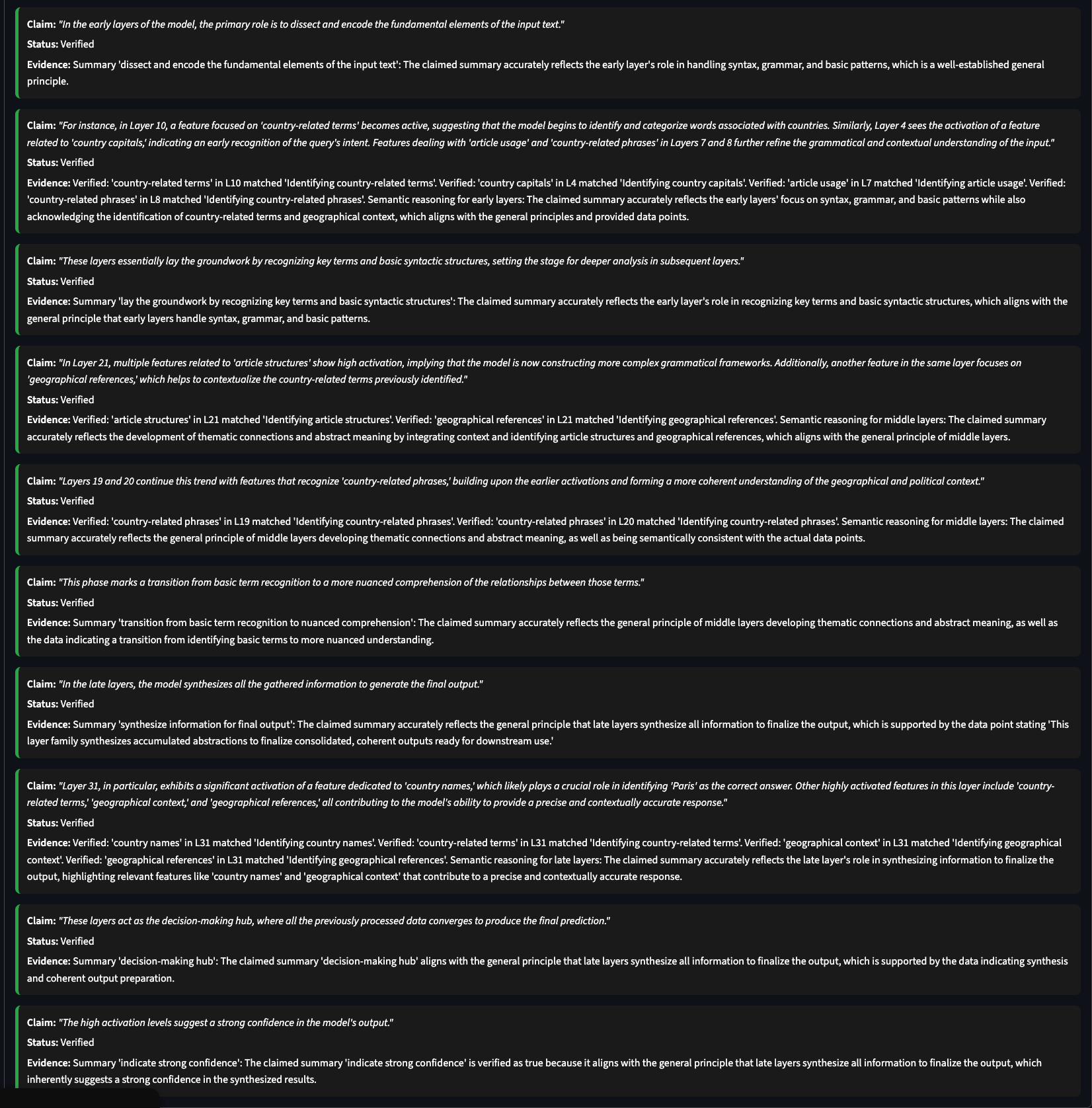}
    \caption{Circuit Trace Analysis: Faithfulness verification of the circuit explanation.}
    \label{fig:app_circuit_faithfulness}
\end{figure*}

\begin{figure*}[h]
    \centering
    \includegraphics[width=\textwidth]{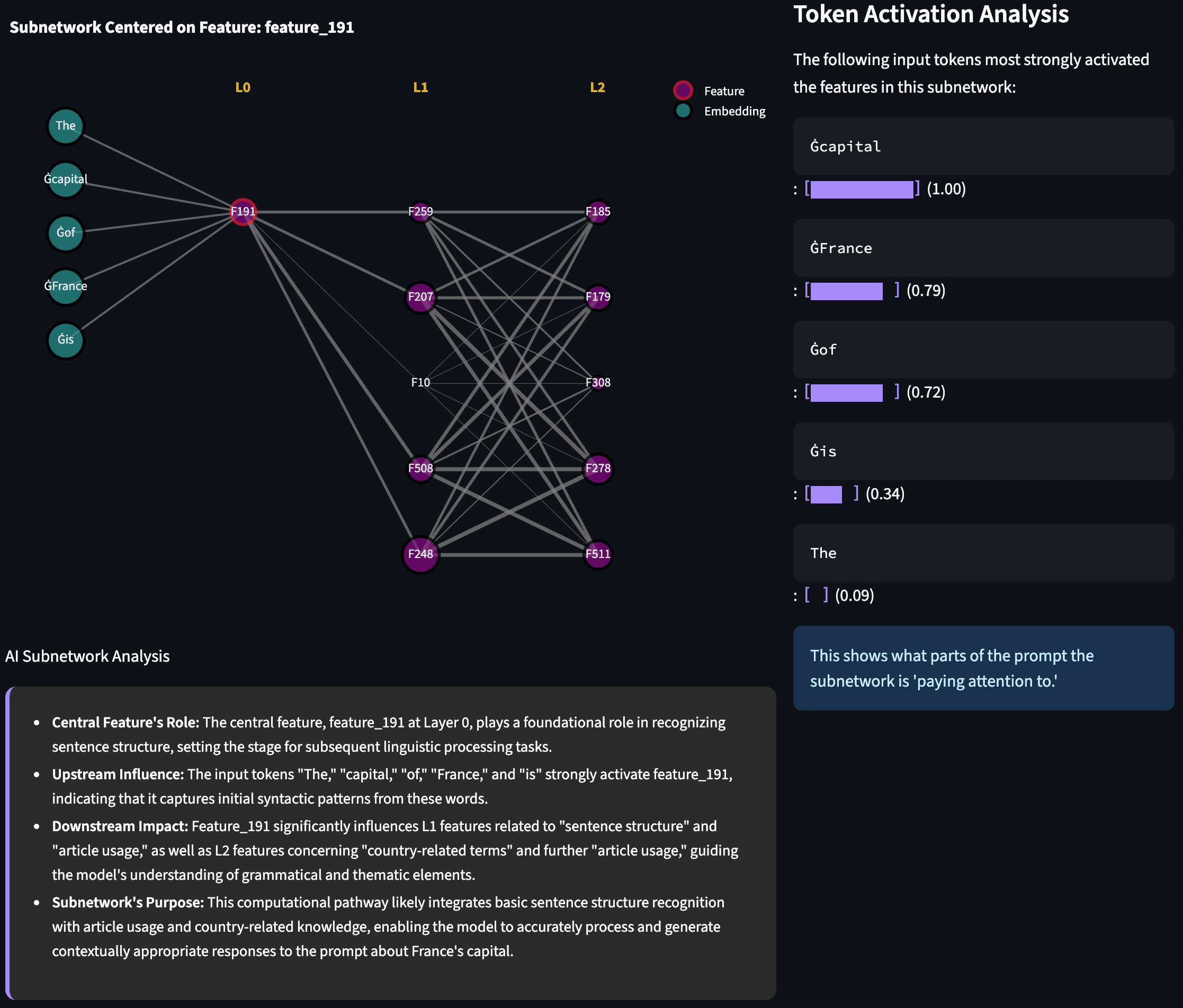}
    \caption{Circuit Trace Analysis: Subnetwork Explorer. This interactive tool allows users to isolate and visualize the specific computational pathways connected to a chosen feature, revealing its upstream influences and downstream effects.}
    \label{fig:app_subnetwork_explorer}
\end{figure*}

\begin{figure*}[h]
    \centering
    \includegraphics[width=\textwidth]{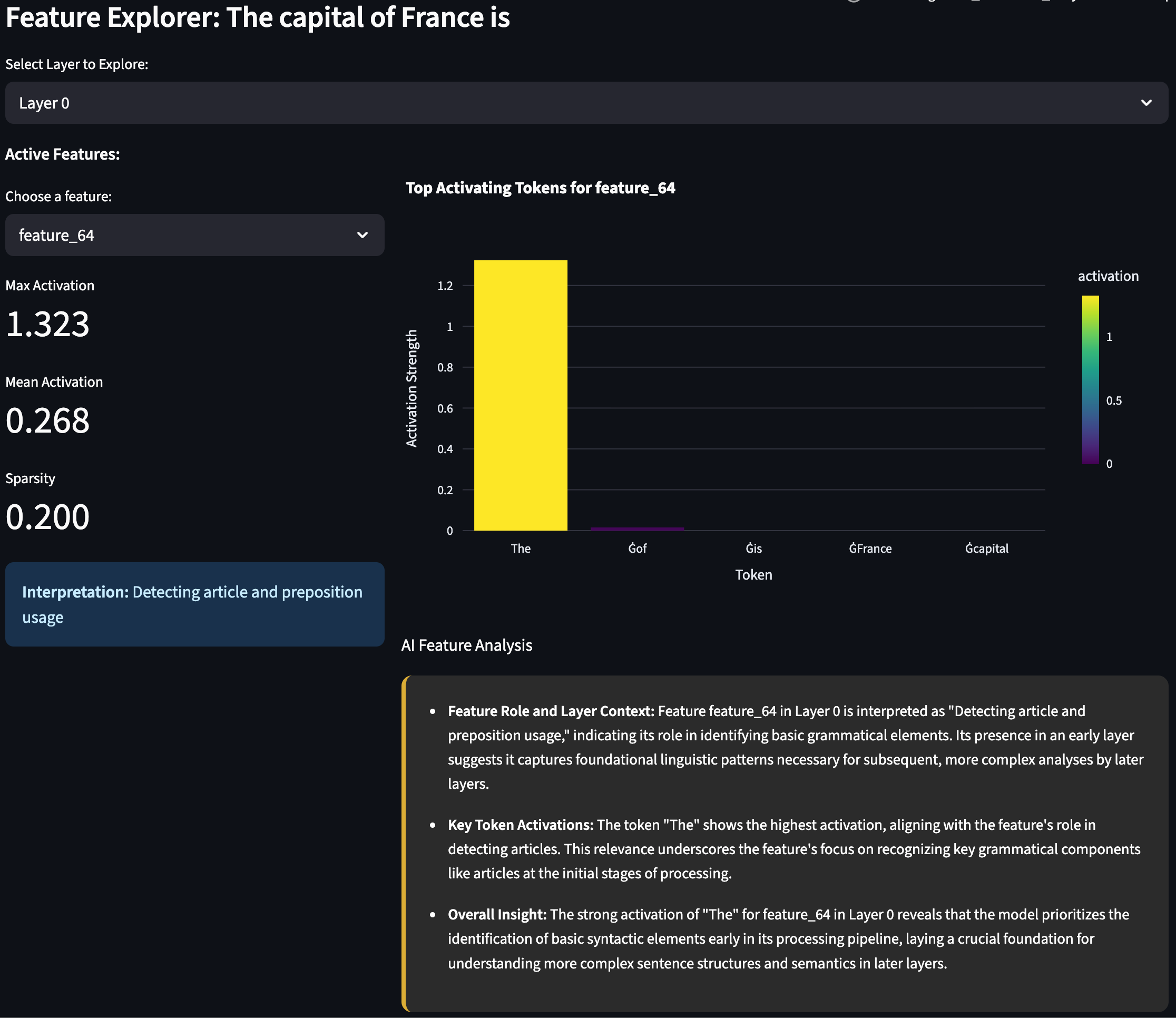}
    \caption{Circuit Trace Analysis: Feature Explorer. Users can inspect individual features in detail, viewing their top activating tokens, sparsity statistics, and AI-generated functional interpretations.}
    \label{fig:app_feature_explorer}
\end{figure*}

\end{document}